\documentclass[journal,onecolumn,11pt]{IEEEtran}
\usepackage{amsmath,epsfig}
\usepackage{setspace}
\usepackage{xcolor}
\usepackage{booktabs}
\usepackage{float}

\begin{document}

\title{Confidence-Guided Semi-supervised Learning in Land Cover Classification}

\author{Wanli Ma, Oktay Karakuş, Paul L. Rosin
\thanks{Wanli Ma, Oktay Karakuş, Paul L. Rosin are with the School of Computer Science and Informatics, Cardiff University, Cardiff CF24 4AG, UK (email: \{maw13, karakuso, rosinpl\}@cardiff.ac.uk).}
}

%
%
%

%
\maketitle
\begin{abstract}
Semi-supervised learning has been well developed to help reduce the cost of manual labelling by exploiting a large quantity of unlabelled data. Especially in the application of land cover classification, pixel-level manual labelling in large-scale imagery is labour-intensive, time-consuming and expensive. However, existing semi-supervised learning methods pay limited attention to the quality of pseudo-labels during training even though the quality of training data is one of the critical factors determining network performance. In order to fill this gap, we develop a \textit{confidence-guided semi-supervised learning} (CGSSL) approach to make use of high-confidence pseudo labels and reduce the negative effect of low-confidence ones for land cover classification. Meanwhile, the proposed semi-supervised learning approach uses multiple network architectures to increase the diversity of pseudo labels. The proposed semi-supervised learning approach significantly improves the performance of land cover classification compared to the classic semi-supervised learning methods and even outperforms fully supervised learning with a complete set of labelled imagery of the benchmark Potsdam land cover dataset.
\end{abstract}

\IEEEpeerreviewmaketitle
\onehalfspacing

\section{Introduction}
\label{sec:intro}
In recent years, with the great success of deep learning in computer vision, automated land cover classification approaches have been significantly improved by using deep learning. Nowadays, the majority of deep-learning land cover classification methods are based on supervised learning \cite{vali2020deep}, which generally requires enormous annotated datasets. Although there are many well-annotated datasets in the computer vision area, it is difficult to generalize deep learning models trained by those datasets to the remote sensing domain. Meanwhile, manual annotation by experts of large-scale remote sensing data, such as satellite products and images captured by drones in complex terrain, is labour-intensive and expensive. Fortunately, a large amount of unlabelled remote sensing data is freely available. Thus, exploring semi-supervised learning approaches for remote sensing applications has become a feasible way to solve the problem of the lack of labelled data \cite{wang2022semi}.

In computer vision, semi-supervised learning just uses a small amount of labelled data along with unlabelled data to eliminate the labour-intensive and expensive annotation stage. This can still lead to competitive performance compared to supervised learning in applications such as image classification \cite{zhang2021flexmatch} and semantic segmentation \cite{hu2021semi}. Specifically, to supervise the deep learning models, mainstream semi-supervised learning methods exploit ``\emph{pseudo}" labels that are fake and/or auto-generated label information obtained as a result of an unsupervised step. The accuracy of pseudo labels is regarded as a key factor affecting the performance of semi-supervised learning approaches. 


This work utilizes an information theoretical approach that evaluates the confidence of pseudo labels to reweight the loss and effectively leverages high-accuracy pseudo labels whilst mitigating the impact of low-accuracy ones. On the other hand, three distinct neural network models are employed in parallel to enhance the diversity of the intermediate network outputs which helps measure the pseudo labels’ confidence. Specifically, the contributions of this work are as follows:
(1) We propose a confidence-guided cross-entropy loss for semi-supervised land cover classification. This is flexible and can be easily transferred to other vision tasks such as semantic segmentation for natural images, and medical image segmentation. 
(2) An adaptive mechanism is designed to adjust the decision criterion automatically (no-interaction setting) for judging the quality of the pseudo labels based on their confidence. 
(3) We promote the investigation of multiple network outputs in terms of an information theory aspect -- entropy -- to weight confidence levels of pseudo labels from each network. Then, we use this information to optimise the unsupervised training process in semi-supervised land cover classification.

\begin{figure*}[t]
\centering
\includegraphics[width=1\textwidth ]{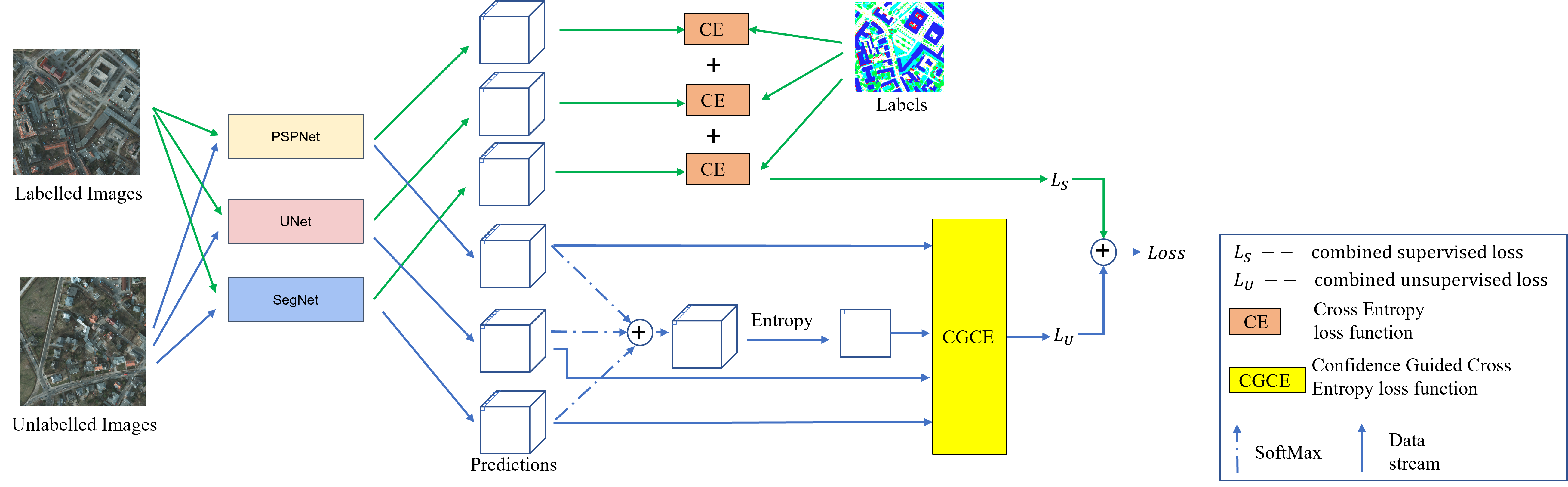}
\caption{Overall framework of the confidence guided semi-supervised learning (CGSSL) approach}
\vspace{-0.3cm}
\label{framework}
\end{figure*}

\section{Background \& Related Work}
\label{sec:related_work}
In general terms, semi-supervised learning is defined as an approach that lies between supervised and unsupervised learning. During the supervised learning step, various widely applied semantic segmentation methods can be used such as PSPNet \cite{zhao2017pyramid}, UNet \cite{ronneberger2015u}, SegNet \cite{badrinarayanan2017segnet}, DeepLabV3+\cite{chen2018encoder}. In current semi-supervised learning research \cite{hu2021semi, wang2022semi_u2pl, xu2022semi} within the field of computer vision, the commonly used network is DeepLabV3+ with a pre-trained backbone e.g. ResNet 50. However, in remote sensing, no specific architecture dominates. Especially when the amount of labelled data is small, due to the exploitation of low- and high-level features via efficient skip-connections, a simpler method like U-net shows competitive (even better) results compared to other classic semantic segmentation networks \cite{zheng2022semi}.

Consistency regularization \cite{french2019semi} describes a class of unsupervised learning algorithms as a part of semi-supervised learning, that are easy to implement and widely compatible with supervised learning segmentation networks. The key idea of consistency regularization is to force perturbed models (or perturbed inputs) to have consistent outputs for unlabelled inputs. Based on this concept, cross pseudo supervision (CPS) \cite{chen2021semi} and CPS-to-$n$-networks (n-CPS) \cite{filipiak2021n} show considerable success, which yields state-of-the-art on semantic segmentation benchmark datasets, e.g. Cityscapes.  However, CPS and n-CPS use pseudo labels to supervise the network regardless of their quality. In addition, perturbed models in those methods have the same structure which causes these networks to tend to output similar predictions. In order to increase the diversity of pseudo labels in parallel, using different segmentation networks stands out to be an efficient and accurate alternative \cite{luo2022semi}. In previous land cover mapping work \cite{ma2022amm}, we have shown that a dual-encoder structure is helpful for leveraging minimal supervision, yet the performance drops significantly when the amount of training data is reduced.

\section{Method}\label{sec:Method}
The proposed semi-supervised learning approach, illustrated in Figure \ref{framework}, uses both labelled and unlabelled data as input into the three different networks (PSPNet \cite{zhao2017pyramid}, UNet \cite{ronneberger2015u}, SegNet \cite{badrinarayanan2017segnet}) in each training iteration. The labelled data is used in a regular supervised learning manner to train these models by using the cross-entropy loss function. In addition, unlabelled data is used to generate pseudo labels, which are exploited to inform each network. The outputs from three networks are added linearly after a softmax layer to generate a comprehensive probability distribution of the classes for all pixels of the input image. In this case, if the probability distributions among the three networks are high-peak unimodal whilst the classifications corresponding to each peak are consistent, the operation of linear addition will keep this unimodal distribution (low uncertainty). Otherwise, e.g. the distributions are not unimodal or they are diverse, the combined prediction will not have a distinct strong peak (high uncertainty). Considering the fact that the information entropy is a measure of uncertainty, we calculated the entropy of the classification distribution based on the combined prediction to investigate the quality of predicted pseudo labels. Furthermore, the proposed confidence-guided cross-entropy loss function is designed to limit the negative contribution of the pseudo labels with high entropy (high uncertainty) to the network parameter optimisation. 

\begin{figure*}[ht]
\centering
\includegraphics[width=1\linewidth ]{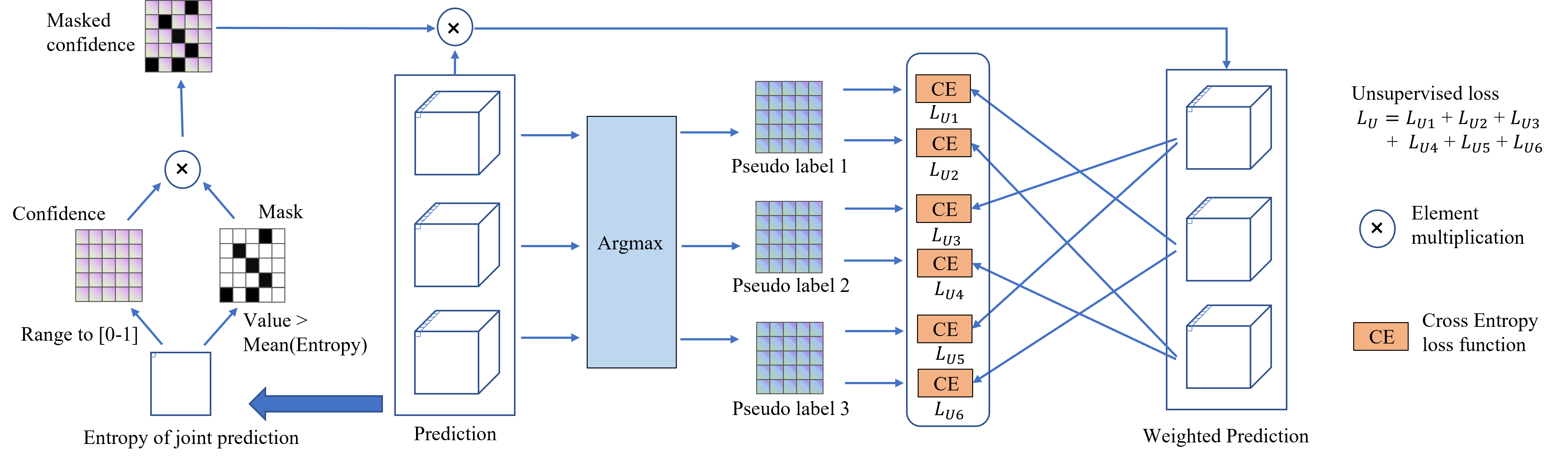}
\vspace{-0.5cm}
\caption{The details of the Confidence-Guided Cross Entropy (CGCE) module.}
\label{CGCE}
\end{figure*}

As shown in Figure \ref{CGCE}, the proposed confidence-guided cross-entropy loss module (CGCE) is used to calculate the unsupervised loss. The aim of this loss is to make use of the high-reliable confidence of predictions to re-weight the standard cross entropy loss at the pixel level based on their entropy among classes. The mean value of entropy is regarded as a threshold to decide on the reliability of the estimated confidence. The unreliable confidence values are assumed to provide limited useful information for re-weighting the loss. Thus, the loss of these pixels is not reweighted and just used in the standard cross-entropy loss function. However, the confidence of predictions above the mean value is regarded as reliable and is used for entropy calculation to re-weight the loss. Instead of directly using the probability of predictions to weight the loss (focal loss \cite{lin2017focal}), entropy is used to form the weight, which represents the confidence of pseudo labels generated from multiple distinct networks. Specifically, the weight $w$ is defined as follows
$w=\frac{\max (\mathcal{I}) - \mathcal{I}}{\max (\mathcal{I})-\min (\mathcal{I})} + 1$,
where $\mathcal{I}$ refers to the entropy of class probability for each pixel.
Thus, since $w \ge 1$ the effect of these pixels is increased during training compared to the pixels with unreliable confidence values. Then, the weight, $w$, is added as a factor to standard cross entropy loss $\ell(x, y)$ to favour the high-quality pseudo labels. 
\begin{equation}
\ell(x, y)= \frac{\sum_{n=1}^N -w \log \frac{\exp \left(x_{n, y_n}\right)}{\sum_{c=1}^C \exp \left(x_{n, c}\right)}}{N}, 
\end{equation}
where $x$ represents the input, $y$ denotes the target class, $w$ signifies the weight, $C$ indicates the number of classes, and $N$ is the batch size. Inspired by \cite{chen2021semi} and \cite{luo2022semi}, the unsupervised loss is acquired by cross-supervision between predictions from different networks. Finally, the total loss $\mathcal{L}$ is set to a linear combination of supervised loss ${L}_{s}$ and unsupervised loss $\mathcal{L}_{u}$ as
\begin{equation}
\mathcal{L}=\mathcal{L}_{s} + \lambda \mathcal{L}_{u},
\end{equation}
where $\lambda$ is the trade-off weight between supervised and unsupervised losses. It is worth noting that the unsupervised loss $\mathcal{L}_{u}$ is the linear addition of 6 losses that result from 3-model cross-supervision, since the pseudo label from each network can supervise the other two networks and leads to two losses, as shown in Figure \ref{CGCE}.

\section{Experiments and Results}
\label{sec:Results}

We evaluated our method using the ISPRS Potsdam dataset \cite{rottensteiner2012isprs}, which consists of 38 multi-source $6000 \times 6000$ patches, including infrared, red, green, and blue orthorectified optical images, and corresponding digital surface models (DSM). We divided these data tiles into $512\times512$ patches, resulting in 3456 training samples and 2016 test samples. Both true orthophoto and DSM modalities have a 5 cm ground sampling distance. The dataset contains six manually classified land cover classes: \textit{impervious surfaces, buildings, low vegetation, trees, cars,} and \textit{clutter/background}. 

In order to compare the proposed method -- CGSSL -- we utilised two classic semi-supervised models of Mean Teacher \cite{tarvainen2017mean} and CPS \cite{chen2021semi}. The quantity of labelled data used in the aforementioned semi-supervised learning approaches is only half (1728 samples) of the whole training split of the Potsdam dataset. We remove the labels of the remaining half and just used the images in the unsupervised part. We also provide the performance of UNet \cite{ronneberger2015u} using fully supervised learning for both the whole and half of labelled data which are named U-Net1 and U-Net2 in the sequel, respectively. The same test set is used to evaluate all models. Thus, when applying the proposed method in real-world scenarios, only a subset of the images would need labelling, and the remainder would be left unlabelled to reduce manual labour.

\begin{table*}[t]
\renewcommand{\arraystretch}{0.7}
  \centering
  \vspace{-0cm}
  \caption{Performance comparison of different methods for Potsdam dataset.}
    \begin{tabular}{p{3cm}p{3cm}p{1.75cm}p{1.75cm}p{1.75cm}p{1.75cm}p{1.75cm}}
    \toprule
          Model & Type & Accuracy & Precision & Recall & mIoU  & $F_1$-score    \\
    \midrule      
    U-Net1$^\dagger$ \cite{ronneberger2015u} & Supervised & 85.36\%	&76.75\%	&81.23\%	&67.59\%	&78.92\% \\
    U-Net2$^*$ \cite{ronneberger2015u}  & Supervised & 84.26\%	&	76.45\%	&	79.32\%	&	66.49\%	&	77.86\%	 \\
    Mean Teacher \cite{tarvainen2017mean} & Semi-Supervised &84.58\%	&	78.52\%	&	80.88\%	&	68.24\%	&	79.68\%	  \\
    CPS \cite{chen2021semi} & Semi-Supervised &85.30\%	&	77.94\%	&	80.75\%	&	68.38\%	&	79.32\%	  \\\midrule 
    CGSSL (ours)  & Semi-Supervised & \textcolor[rgb]{1,0,0}{\textbf{86.59\%}}
 & 	\textcolor[rgb]{1,0,0}{\textbf{79.06\%}}&	\textcolor[rgb]{1,0,0}{\textbf{83.54\%}} &	\textcolor[rgb]{1,0,0}{\textbf{70.17\%}} &	\textcolor[rgb]{1,0,0}{\textbf{81.24\%}}  \\
    \bottomrule
    \end{tabular}\\
        \small{\textit{$^\dagger$U-Net1 was trained with the whole 3456 labelled samples. $^*$U-Net2 was trained with 1728 labelled samples.}}
  \label{tab:Performance}%
  \vspace{-0.3cm}
\end{table*}%

\begin{figure*}[ht]
\centering
\includegraphics[width=1\linewidth ]{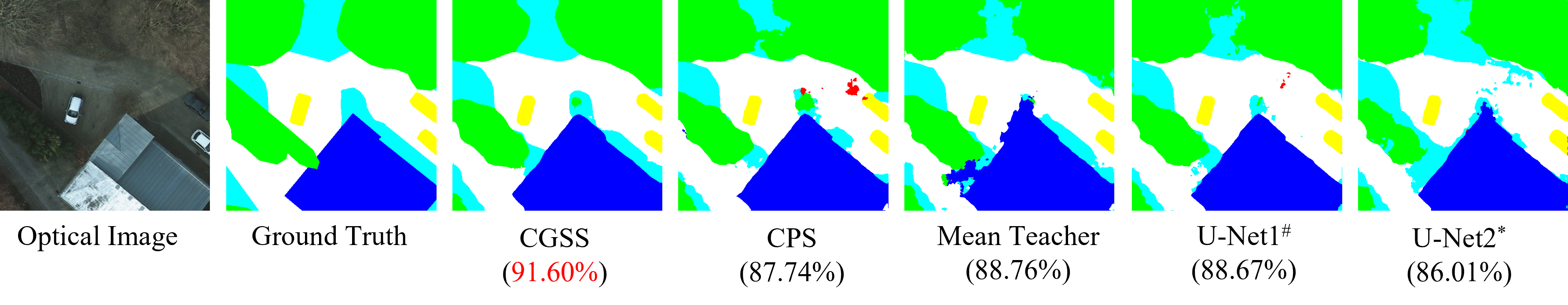}
\vspace{-0.8cm}
\caption{Visual Results of each method on Potsdam Dataset. Values in parentheses refer
to percentage accuracy. $^\#$U-Net1 was trained with the whole 3456 labelled samples. $^*$U-Net2 was trained with 1728 labelled samples.
}
\label{Visual_results}
\vspace{-0.3cm}
\end{figure*}

Our experiments were implemented by Pytorch. We used a mini-batch SGD optimizer that adopted a polynomial learning rate policy. All the experiments were performed on an NVIDIA A100-sxm in a GW4 Isambard. We thoroughly evaluated all models using class-related performance metrics, including accuracy, precision, recall, mean intersection over union (mIoU), and $F_1$-score. As shown in Table \ref{tab:Performance}, CGSSL shows the best performance in terms of all performance metrics. In particular, CGSSL improves recall significantly due to the great reduction of false negatives in prediction. Even though CGSSL only uses half of the labelled data, its performance is even better than UNet1 which is trained with the whole dataset. Figure \ref{Visual_results} shows an example of predictions for all methods where CGSSL is mostly close to the ground truth.

\section{Conclusion}
\label{sec:Conclusion}
In this paper, we introduced an innovative semi-supervised learning approach for land cover classification that utilizes a confidence-guided cross-entropy loss. In particular, an adaptive loss was provided for semi-supervised learning to exploit high-quality pseudo labels and limit the effect of low-quality pseudo labels with an information theory perspective. Our approach is also flexible and can be transferred to various other semi-supervised learning tasks. The proposed method shows considerable performance for land cover classification, and benefits from unlabeled data. Meanwhile, since three networks are required to increase the diversity of pseudo labels in training processing, one of the drawbacks of this method is the increased computational requirement, which means that it might not be appropriate for edge computing devices in practical applications. Thus, our future work aims to further develop computationally cheaper segmentation architectures for semi-supervised learning. 

\begingroup
\bibliographystyle{IEEEtran}
\bibliography{refs}
\endgroup

\end{document}